\documentclass[10pt,twocolumn,letterpaper]{article}

\usepackage[pagenumbers]{cvpr} 

\usepackage[dvipsnames, table]{xcolor}

\definecolor{cvprblue}{rgb}{0.21,0.49,0.74}
\usepackage[pagebackref,breaklinks,colorlinks,citecolor=cvprblue]{hyperref}
\usepackage{multirow}

\title{A Unified Framework for 3D Point Cloud Visual Grounding}

\author{
    Haojia Lin$^{1}$,
    Yongdong Luo$^{1}$,    
    Xiawu Zheng$^{2}$,
    Lijiang Li$^{1}$,
    Fei Chao$^{1}$,
    Taisong Jin$^{1}$, 
    Donghao Luo, \\
    Yan Wang,
    Liujuan Cao$^{1}$,    
    Rongrong Ji$^{1}$ \\
    {\tt\small $^1$Xiamen University. 
    $^2$Peng Cheng Laboratory. }
}
\begin{document}
\maketitle

\begin{abstract}
Thanks to its precise spatial referencing, 3D point cloud visual grounding is essential for deep understanding and dynamic interaction in 3D environments, encompassing  3D Referring Expression Comprehension (3DREC) and Segmentation (3DRES).
We argue that 3DREC and 3DRES should be unified in one framework, which is also a natural progression in the community. 
To explain, 3DREC help 3DRES locate the referent, while 3DRES also facilitate 3DREC via more fine-grained language-visual alignment. 
To achieve this, this paper takes the initiative step to integrate  3DREC and 3DRES into a unified framework, termed 3D Referring Transformer (3DRefTR). 
Its key idea is to build upon a mature 3DREC model and leverage ready query embeddings and visual tokens from the 3DREC model to construct a dedicated mask branch. 
Specially, we propose Superpoint Mask Branch, which serves a dual purpose: i) By harnessing on the inherent association between the superpoints and point cloud, it eliminates the heavy computational overhead on the high-resolution visual features for upsampling; 
ii) By leveraging the heterogeneous CPU-GPU parallelism, while the GPU is occupied generating visual and language tokens, the CPU concurrently produces superpoints, equivalently accomplishing the upsampling computation. 
This elaborate design enables 3DRefTR to achieve both well-performing 3DRES and 3DREC capacities with only a 6\% additional latency compared to the original 3DREC model. 
Empirical evaluations affirm the superiority of 3DRefTR. Specifically, on the ScanRefer dataset, 3DRefTR surpasses the state-of-the-art 3DRES method by 12.43\% in mIoU and improves upon the SOTA 3DREC method by 0.6\% Acc@0.25IoU. 
The codes are at \url{https://github.com/Leon1207/3DRefTR}.
\end{abstract}

\begin{figure}[h!]
  \centering 
  \includegraphics[width=1.0\linewidth]{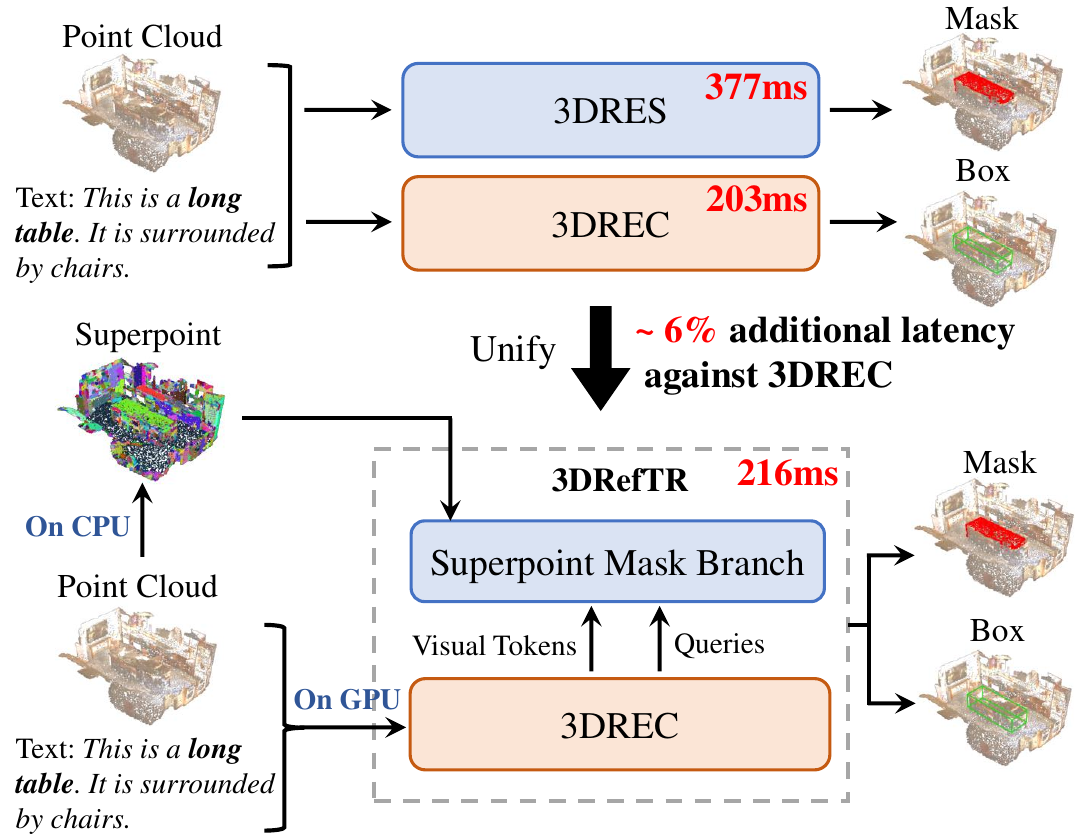}
  \caption{Utilizing two model specialized respectively for 3DREC and 3DRES, the computational cost for executing these two tasks is (203 + 377) milliseconds given the constraints of limited computation resource. 
   In contrast, with Superpoint Mask Branch, 3DRefTR processes superpoints (On CPU) and visual tokens (On GPU) in parallel, unifying 3DREC and 3DRES with only a 6\% increase in total latency compared to the original 3DREC model. An exhaustive efficiency analysis is presented in Sec. \ref{efficiency_analysis}.}
  \label{fig_intro} 
\end{figure}

\section{Introduction}

The task of 3D Visual Grounding (3DVG) \cite{viewrefer,multi3drefer,ns3d,ham,butd,eda,3dvg,sat,instancerefer,languagerefer,3dsps,lamm,tgnn}, has been gaining significant interest as a crucial 3D cross-modal challenge. 
The core goal of 3DVG is to locate objects in 3D point clouds using linguistic queries, which is essential for the development of advanced applications such as assistive robotics and immersive AR/VR. 
Additionally, The task's emphasis on the interplay between language and spatial understanding also makes it a key driver for progress in the 3D field, enhancing AI's spatial cognition.

3DVG can be encountered two divergent yet interrelated tasks: 3D Referring Expression Comprehension (3DREC) \cite{viewrefer,multi3drefer,ns3d,ham,butd,eda,3dvg,sat,instancerefer,languagerefer,3dsps,lamm} and 3D Referring Expression Segmentation (3DRES) \cite{tgnn}. 
Various advances have been made in the domain of 3DREC, several contributions stand out: 
SAT \cite{sat} improves point cloud features extraction by 2D images assistance; 
BUTD \cite{butd} iteratively enhances object localization precision by incorporating Transformer's encoder-decoder architecture; 
EDA \cite{eda} refines object queries through explicit text decoupling and dense alignment.
In contrast, the progress in 3DRES has been modest. This is primarily because segmenting objects from cluttered 3D point clouds according the referring text is not only computationally intensive but also more complex when compared to 3DREC. 
TGNN \cite{tgnn} is the first to paves path to 3DRES, but its reliance on a pre-trained 3D instance segmentation model limits efficiency and precision, highlighting the need for further research in this emerging area.

It is a natural idea to integrate 3DREC and 3DRES for mutual enhancement, as similar to the established practice of multi-task learning in 2D vision \cite{mcn,reftr,seqtr}. 
To be more specific, the point-level supervision provided by 3DRES amplifies the capabilities of 3DREC, ensuring a more coherent vision-language alignment during multimodal training. 
Meanwhile, 3DREC excels at localizing the referent object, addressing the limitations of RES in identifying the correct instance. 
Our ambition is to harmoniously fuse these tasks, proposing a unified methodology for 3DREC and 3DRES. 
In the domain of 2D visual grounding, analogous methods have been developed, offering insight into possible solutions for 3DVG. 
Yet, due to the intrinsic data characteristics of 3D point clouds, these techniques are challenging to transfer directly. 
For example, RefTR \cite{reftr} allocates specific backbone features for each query embedding and undergoes an upsampling decoder. However, due to the vast number of 3D point clouds, adopting this technique will result in an unfeasible computational overhead and memory consumption. 
Another method, SeqTR \cite{seqtr}, aims to obtain masks by predicting polygons that encapsulate the masks, but three-dimensional spaces lack well-defined polygons to encapsulate 3D objects. 

As shown in Fig. \ref{fig_intro}, the fundamental concept behind 3DRefTR entails an extension of a well-established 3DREC model, capitalizing on pre-existing query embeddings and visual tokens derived from the 3DREC model to construct a dedicated mask branch. 
This strategic approach effectively mitigates the additional computational cost that would arise from the integration of 3DRES capability. 
To further alleviate the computational burden associated with feature upsampling within the mask branch, we introduce the incorporation of superpoints \cite{superpoint}. 
This introduces two primary advantages: i) Exploiting the inherent associations between the superpoints and point cloud to eliminate the intensive computation on high-resolution visual features during the upsampling process; ii) Leveraging heterogeneous CPU-GPU parallelism enables simultaneous CPU-driven superpoint generation while the GPU focuses on generating visual tokens, equivalently achieving the upsampling process. 
With the integration of the superpoint mask branch,  3DRefTR  seamlessly unifies the capabilities of 3DREC and 3DRES with only a 6\% addition latency against the original 3DREC model. 
Furthermore, the synergistic effect of unifying these two closely related tasks within a single model leads to mutual enhancement. 

To sum up, the main contributions of this paper are as follows:
\begin{itemize}
    \item \textbf{Efficient Superpoint Mask Branch}:  Observing the heavy computational load of upsampling in 3DRES, we approach the issue through introducing the Superpoint Mask Branch, which shifts the upsampling computation to the CPU while the GPU processes the backbone features, thus enhancing efficiency.
    \item \textbf{Unified Framework for 3D Visual Grounding}: With the proposed superpoint mask branch, this paper further introduces the 3D Referring Transformer (3DRefTR). It is a pioneering framework that concurrently addresses both 3DREC and 3DRES tasks simultaneously.
    3DRefTR delivers enhanced 3DRES and 3DREC capabilities with just a 6\% latency increase compared to the original 3DREC model.
    \item \textbf{Superior performance on both tasks}: Empirical results on the ScanRefer dataset show that 3DRefTR outperforms the leading 3DRES method by a margin of 12.43\% in mIoU and exceeds the state-of-the-art 3DREC method by 0.6\% in Acc@0.25IoU.
\end{itemize}

\begin{figure*}[h!]
  \centering 
  \includegraphics[width=1.0\linewidth]{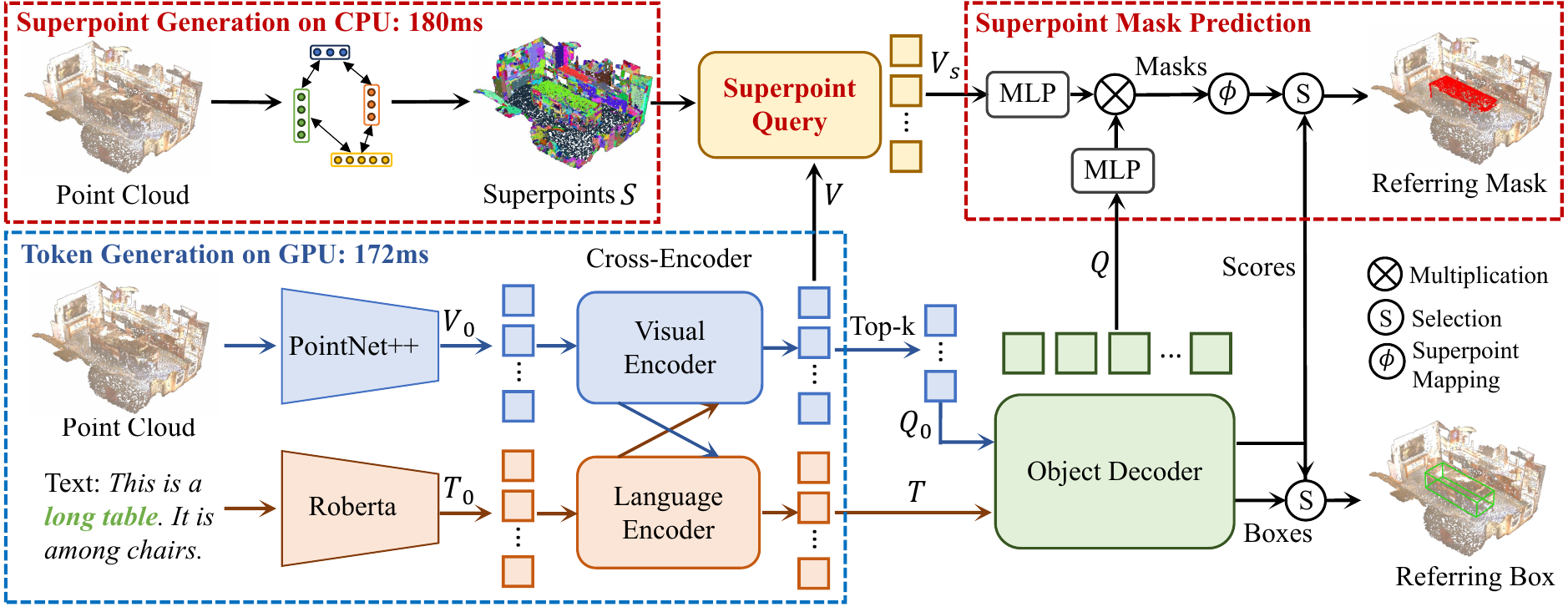}

  \caption{The framework of 3DRefTR builds upon a mature 3DREC model (bottom in the figure) with a SuperPoint Mask Branch. For 3DRES, after obtaining the superpoints $S$ and visual tokens $V$ of each scene in parallel, $S$ is used as the key to retrieve the corresponding nearest-neighbor features $V_s$ from $V$ by Superpoint Query. $V_s$ and query embeddings $Q$ are multiplied  after MLP projection to obtain the candidate masks, which are then mapped to the original resolution via superpoint mapping $\phi$. Finally, the referring mask is selected from the candidate masks via the referring scores. }
  \label{fig_framework} 
\end{figure*}
\section{Related Work}
\subsection{3D Visual Grouding}
3D Visual Grounding (3DVG) aims to identify the targeted object in a 3D scene based on linguistic cues. 
Broadly, 3DVG encompasses two pivotal tasks: 3DREC and 3DRES. 
Within the 3DREC realm, most recent works \cite{3dvg,sat} adopt a two-stage pipeline. Initially, they harness ground truth or a 3D object detector \cite{pointgroup,groupfree} to produce the object proposals. Subsequently, these methods use text and 3D encoder \cite{roberta,pointnet2} to extract features and then ground the target one after the feature fusion. 
For instance, InstanceRefer \cite{instancerefer} ingeniously construes the task as instance-matching, 
while LanguageRefer \cite{languagerefer} transforms it into a linguistic modeling task by replacing the 3D features with predicted object labels. 
BUTD \cite{butd} iteratively enhances object localization precision by incorporating Transformer's encoder-decoder architecture\cite{butd}. 
3D-SPS \cite{3dsps} first proposes a single-stage network, conceiving 3DVG as a keypoint selection problem.
In contrast, the domain of 3DRES remains relatively untouched, with TGNN \cite{tgnn} being the solitary pioneer. This work hinges on an instance segmentation model, utilizing text-guided Graph Neural Networks to identify the text-referred instance center.

A clear gap in current research is the lack of studies that tackle both 3DREC and 3DRES together. 
Our study aims to fill this gap by combining these two tasks into one unified framework, which not only simplifies the model deployment but also allows the tasks to benefit from each other.

\subsection{3D Detection and Segmentation} 
3D detection \cite{votenet,groupfree,tokenfusion,cagroup3d,fcaf3d,3ddetr} and segmentation \cite{spformer,mask3d,sstnet,softgroup,isbnet,queryrefine} algorithms stand as the cornerstone for 3DVG. 
Their significance in advancing 3DVG methods manifests primarily in two aspects. First, these detection algorithms directly provide object proposals for two-stage 3DVG methodologies as a part of the pipeline. 
Second, recent state-of-the-art 3DVG approaches, such as BUTD \cite{butd} and EDA \cite{eda}, have incorporated the query selection modules from Groupfree \cite{groupfree}, indicating a converging trend of these technologies. 
Parallel to the latest 3DREC models like \cite{butd,eda}, recent 3D instance segmentation models, namely SPformer \cite{spformer} and mask3D \cite{mask3d}, also adopt the DETR-based architecture \cite{detr}. 
However, due to the adoption of parametric query embeddings in these methods, their localization capabilities are inferior compared to the detectors that utilize feature-based query embeddings \cite{groupfree}. 
In this context, our strategy in this paper is to leverage the precise localization capacity of the query embeddings in 3DREC. This approach not only mitigates the overhead introduced by incorporating 3DRES but also ensures the 3DREC capability remains uncompromised.


\section{Method} 
The key idea of 3DRefTR is to harness the ready localization capability of 3DREC model to develop the 3DRES potentials, subsequently utilizing the point-level supervisory signals of 3DRES to augment 3DREC performance. 
The architecture of the adopted 3DREC model \cite{eda} is illustrated in Sec. \ref{preliminary}. 
To develop the 3DRES capacity, an high-resolution mask branch is appended to the 3DREC model for mask prediction, which is detailed in Sec. \ref{hr_branch}. 
However, recognizing the computational intensity of upsamping in the mask branch, we propose an innovative superpoint mask branch, as described in Sec. \ref{superpoint_branch}. 
This novel branch significantly reduces computational load, as we demonstrate in the efficiency analysis provided in Sec. \ref{efficiency_analysis}. 
Last, the training scheme and the inference process is illustrated in Sec. \ref{training} and Sec. \ref{inference}, respectively.

\subsection{Preliminaries: 3DREC Model}\label{preliminary}

As shown in Fig. \ref{fig_framework}, the 3DREC model (bottom in the figure) is a typical DETR-like \cite{detr} model, which is composed of a backbone, a cross-encoder, and an object decoder. 

\textbf{Visual and Textual Backbone. }
PointNet++ \cite{pointnet2} produces the visual tokens $V_0 \in \mathbb{R}^{n \times d}$ from the input point cloud $P\in \mathbb{R}^{N \times 3}$, while Roberta-base \cite{roberta} processes referential text for text tokens $T_0 \in \mathbb{R}^{l \times d}$, where $n$ denotes the number of visual token, $l$ denotes the number of text token, $N$ denotes the point cloud size, and $d$ is the feature dimension. 

\textbf{Cross-Encoder. }
The visual tokens $V_0 $ and text tokens $T_0$ enter a dual-pathway cross-encoder alternatively stacking with self-attention, cross-attention, and FFN layers, producing thoroughly fused features $V \in \mathbb{R}^{n \times d}$ and $T \in \mathbb{R}^{l \times d}$. 
The output visual tokens $V \in \mathbb{R}^{n \times d}$ is input to the KPS module proposed by \cite{groupfree}. 
The top-k object queries $Q_0 \in \mathbb{R}^{k \times d}$ are chosen and then fed into the object decoder and the query scoring branch.  

\textbf{Object Decoder. }
The top-k object queries $Q_0$ and the text tokens $T$ enter the object decoder which consists of stacked Transformer decoder layers, producing the final query embeddings $Q \in \mathbb{R}^{k \times d}$.
On top of these object query embeddings $Q$, the decoder has two branches respectively for box regression and box-text alignment. 
The box branch dynamically updates boxes $B\in \mathbb{R}^{k \times 6}$ and provides position embeddings for query embeddings in each Transformer decoder layer. 
The alignment branch output the referring scores $s\in \mathbb{R}^{k}$ for the query embeddings to determine which query best matches the referent. 
Given the existing box and alignment branches, the query embeddings $Q$ in 3DREC model exhibit inherent localization abilities. 
The pivot is to harness these embeddings to construct the mask branch to unlock the segmentation potential of the model. 

\subsection{High-Resolution Mask Branch}\label{hr_branch}
Inspired by MaskDINO \cite{maskdino}, the most straightforward approach to construct the mask branch is to first pass the visual tokens through an upsampling decoder $U(\cdot)$, obtaining the original-resolution visual tokens $V_0 \in \mathbb{R}^{N \times d}$. 
Subsequently, by multiplying it with the ready query embeddings $Q$ in the 3DREC model, we derive a mask prediction $M_0 \in [0, 1]^{N \times k}$ with the original resolution. 
Employing this mask branch construction strategy, we can realize a unified model proficient in both 3DREC and 3DRES tasks, which is named 3DRefTR-HR. 

However, a challenge arises due to the vast original point cloud size $N$, which often exceeds 50,000 in datasets like ScanNet. 
This addition of the mask branch subsequently inflates the computational demand for the entire model by approximately 140\% (As shown in Tab. \ref{tab_latency}). 
Such a surge in computation, when taken in the broader context of real-time applications and resource constraints, renders this naive approach suboptimal. 
It thus encourages us to search for a more efficient mask branch design. 

\subsection{Superpoint Mask Branch}\label{superpoint_branch}
In this subsection, our objective is to develop an approach that strikes a balance between performance and efficiency for the added mask branch in our 3DRefTR. 
To this end, we turn to superpoints \cite{superpoint} as a bridge connecting low-resolution features with the high-resolution point cloud, paving the way for a more efficient upsampling technique. 
The model with this upsampling technique is named as 3DRefTR-SP. 

\textbf{Superpoint Generation on CPU. }
For the input point cloud $P\in \mathbb{R}^{N \times 3}$, superpoints $S\in \mathbb{R}^{m \times 3}$ define geometrically homogeneous neighborhoods of its local points, and are usually computed by over-segmenting $P$ using graph partition \cite{sstnet}. 
In another word, there is a mapping $\phi: S\in \mathbb{R}^{m \times 3} \rightarrow P\in \mathbb{R}^{N \times 3}$ between the superpoints $S$ and the orginal point cloud $P$. 
Therefore, the prediction of the original mask $M_0$ can be reformulated as the prediction of the superpoint mask $M_s$.  
Because the $\phi$ is equivalent to the upsampling function and the superpoint generation can be conducted on CPU, the computation of upsampling is shifted to precede the cross-encoder output, parrallelling with  producing the visual tokens $V$. 

\textbf{Superpoint Query. }
The next step involves acquiring the visual embedding for each superpoint. 
To achieve this, we design a parameter-free superpoint query module. 
Given the output visual tokens $V'$ from the cross-encoder and the superpoints $S$, we employ the ball query \cite{pointnet2} to find the nearest neighbor tokens in coordinate space for each superpoint. 
Subsequent max-pooling is used for aggregation, producing the superpoint embeddings $V_s \in \mathbb{R}^{m \times d}$.

\textbf{Superpoint Mask Prediction. }
Lastly, the generated superpoint embeddings $V_s$ and the ready query embeddings $Q$ from the 3DREC model are separately passed through a Multi-Layer Perceptron (MLP). 
Their respective outputs are then multiplied, resulting in the superpoint binary masks $M_s \in [0, 1]^{m \times k}$. 
The candidate masks $M_0 \in [0, 1]^{N \times k}$ with original resolution can be easily acquired from $M_s$ via the superpoint mapping $\phi$. 
Finally, the reffering mask $M_r\in [0, 1]^N$ is selected from $M_0$ via the referring scores $s$. 

\begin{table}[]
\begin{tabular}{l|ccc}
\hline
\multicolumn{1}{c|}{Method} & \multicolumn{1}{c}{\begin{tabular}[c]{@{}c@{}}FLOPs\\ (G)\end{tabular}} & \multicolumn{1}{c}{\begin{tabular}[c]{@{}c@{}}Component Time\\ (ms)\end{tabular}} & \multicolumn{1}{c}{\begin{tabular}[c]{@{}c@{}}Total\\ (ms)\end{tabular}} \\ \hline
\multirow{2}{*}{EDA \cite{eda}}        & \multirow{2}{*}{9.8}    & Before $V$ (GPU):172   & \multirow{2}{*}{203} \\
                            &                         & After $V$ (GPU):31     &            \\ \hline
\multirow{2}{*}{3DRefTR-HR} & \multirow{2}{*}{24.38}  & Before $V$ (GPU):172   & \multirow{2}{*}{405} \\
                            &                         & After $V$ (GPU):233     &            \\ \hline
\multirow{3}{*}{3DRefTR-SP} & \multirow{3}{*}{10.68}  & Before $S$ (CPU):180   & \multirow{3}{*}{216} \\
                            &                         & Before $V$ (GPU):172   &                      \\
                            &                         & After $V$ (GPU):36     &               \\ \hline
\end{tabular}
    \caption{Efficiency analysis for 3DRefTR-SP. $V$ denote the visual tokens produced by the backbone and $S$ denotes the generated superpoints, which can be found in Fig. \ref{fig_framework}. }
    \label{tab_latency}
\end{table}
\subsection{Highlights on Efficiency} \label{efficiency_analysis}
The essence of Superpoint Mask Branch is its dual-edge advantage: i) it eliminates the heavy computational overhead on the high-resolution visual features during upsampling; ii) by conducting upsampling computations prior to the cross-encoder output, the approach leverages heterogeneous CPU-GPU parallelism, achieving upsampling with minimal overhead.
To illustrate the advantage of Superpoint Mask Branch, the total model latency and the components latency are measured as shown in Tab. \ref{tab_latency}. 

\begin{table*}[t!]
  \centering
\setlength{\tabcolsep}{5.0mm}  
\renewcommand{\arraystretch}{1.05} 
\begin{tabular}{l|cc|cc|cc|c}
\hline
\multicolumn{1}{c|}{} & \multicolumn{2}{c|}{Unique(19\%)} & \multicolumn{2}{c|}{Multiple(81\%)} & \multicolumn{2}{c|}{Overall} &  \\
\multicolumn{1}{c|}{\multirow{-2}{*}{Method}}                        & 0.25             & 0.5            & 0.25              & 0.5             & 0.25          & 0.5    & {\multirow{-2}{*}{mIoU}}     \\ \hline
TGNN \cite{tgnn}   &         -         &          -      &          -         &         -        &        37.50       &      31.40       &    27.80  \\
3DRESTR                    &         78.99         &         54.19       &          40.19         &    22.07             &        45.98       &      26.87       &   28.74   \\ 
EDA-box2mask \cite{eda}  &        84.71          &        56.94        &          49.98         &         37.03        &        55.16       &       39.99      &   35.03   \\
\rowcolor{gray!10}\textbf{3DRefTR-HR (ours)}  &\textbf{89.64} &\textbf{77.03}&\textbf{52.31}&\textbf{43.71}&\textbf{57.88}&\textbf{48.69}&\textbf{41.24}\\
\rowcolor{gray!10}\textbf{3DRefTR-SP (ours)}  &\textbf{87.88} &\textbf{69.77}&\textbf{51.61}&\textbf{41.91}&\textbf{57.02}&\textbf{46.07}&\textbf{40.76}\\ \hline
\multicolumn{7}{l}{\textit{single-stage}} \\ 
\hline
EDA-box2mask \cite{eda}     &           84.50       &        55.46        &        49.02           &         36.38        &       54.31       &       39.23      &   34.67   \\
\rowcolor{gray!10}\textbf{3DRefTR-HR (ours)}  &\textbf{89.71}    &\textbf{76.11}    &\textbf{51.45}      &\textbf{43.07} &\textbf{57.16} &\textbf{48.00} &\textbf{40.75}   \\
\rowcolor{gray!10}\textbf{3DRefTR-SP (ours)}  &\textbf{86.75}    &\textbf{69.56}    &\textbf{50.67}      &\textbf{41.18} &\textbf{56.06} &\textbf{45.36} &\textbf{40.23}   \\  \hline
\end{tabular}
    \caption{The 3D referring expression segmentation results on ScanRefer. Compared to the two-stage methods, TGNN \cite{tgnn} and two other established baselines, the signle-stage 3DRefTR-SP supasses with a margin of 12.43\%, 11.49\%, and 5.2\% in mIoU, respectively. }
    \label{tab_3dres}
\end{table*}
\textbf{Compared with 3DRefTR-HR. }
As depicted in Tab. \ref{tab_latency}, the introduction of an upsampling decoder in 3DRefTR-HR inflates computational demands by 2.5 times compared to the original model and doubles the inference latency. 
In contrast, the FLOPs and latency of the 3DRefTR-SP model see only modest increments of 8\% and 6\% against the 3DREC model,  respectively. 
Of paramount significance is that the Superpoint Mask Branch shifts the computational burden of upsampling (i.e., superpoint generation) to precede the cross-encoder output. 
As shown in Tab. \ref{tab_latency}, the model's latency from input to cross-encoder output $V$ is 172 ms, roughly equivalent to the 180 ms consumed by superpoint $S$  generation. 
Given that superpoint generation is executed on the CPU, it doesn't compete for computational resources with the main model components and can run in parallel, which results in  a much higher efficiency.  

\textbf{Compared with SPformer \cite{spformer}and SSTNet \cite{sstnet}. }
It is essential to highlight the distinctions of superpoint usage between our approach, SPformer, and SSTNet.
These methods employ superpoints as an integral part of their backbone for feature aggregation. 
As a result, the generation of superpoints inevitably consumes a significant portion of the model's forward-pass time. 
However, our approach contrasts this by leveraging the heterogeneous computing of CPU and GPU.
During the time-gap where our mask branch awaits the output visual tokens from the cross-encoder, we allocate the superpoint generation task. 
This procedure effectively exploits CPU-GPU parallelism, facilitating the addition of 3DRES capabilities to the 3DREC model with almost negligible additional cost. 


In summing up, the pivotal insight of this paper is the maximal reuse of a well-performing 3DREC model to seamlessly and efficiently integrate 3DRES capabilities into a unified framework. 
The design of our proposed Superpoint Mask Branch seamlessly aligns with this fundamental principle. 

\subsection{Multi-task Training}\label{training}
This subsection illustrates the loss functions and the joint training scheme for 3DRefTR. 

\textbf{Loss Function.} 
For the 3DREC task, we follow the training scheme of EDA \cite{eda}, which adopts five loss functions for each layer of the object decoder:  
box center coordinate prediction with a smooth-$L_1$ loss $\mathcal{L}_{\text {coord }}$, 
box size prediction with a smooth-$L_1$ loss $\mathcal{L}_{\text {size }}$, 
GIoU loss \cite{giouloss} $\mathcal{L}_{\text {giou }}$, 
the semantic alignment loss $\mathcal{L}_{\text {sem }}$,
and the position alignment loss $\mathcal{L}_{\text {pos }}$. 
The loss of $l$-th decoder layer is the combination of these 5 loss terms by weighted summation: 
$$
\mathcal{L}_{\text {dec }}^{(l)}=\beta_1 \mathcal{L}_{\text {coord }}^{(l)}+\beta_2 \mathcal{L}_{\text {size }}^{(l)}+\beta_3 \mathcal{L}_{\text {giou }}^{(l)}+\beta_4 \mathcal{L}_{\text {sem }}^{(l)}+\beta_5 \mathcal{L}_{\text {pos }}^{(l)}.
$$
The losses on all decoder layers are averaged to form the total 3DREC loss:
$$
\mathcal{L}_{\text {rec }}=\frac{1}{L} \sum_{l=1}^L \mathcal{L}_{\text {dec }}^{(l)}.
$$
For the 3DRES task, we follow RefTR \cite{reftr} and apply the focal loss \cite{focalloss} $\mathcal{L}_{\text {focal }}$ 
and the dice loss \cite{diceloss} $\mathcal{L}_{\text {dice }}$ on the top of the last layer of the object decoder. 
Together with the KPS loss \cite{groupfree} $\mathcal{L}_{\text {kps }}$ for the query selection, the final loss for 3DRefTR is as follows: 
$$
\mathcal{L}=\alpha_1\mathcal{L}_{\text {rec}}+\alpha_2\mathcal{L}_{\text {focal}}+\alpha_3\mathcal{L}_{\text {dice}}+\alpha_4\mathcal{L}_{\text {kps }}.
$$

\textbf{Joint Training.} 
For efficient convergence, we initiate the joint training process by loading pre-trained weights from the 3DREC model. 
Subsequently, the training unfolds with a learning rate set at 0.01 times the standard rate for the 3DREC component, while the superpoint mask branch is trained using the standard learning rate. 
This training scheme has demonstrated its efficacy in maintaining a favorable equilibrium between the performance of 3DREC and 3DRES. 
A thorough analysis of the impact of this training scheme will be shown in the section of the Ablation Study.

\subsection{Inference}\label{inference}
During inference, given an input point cloud and a referring text, 3DRefTR outputs the superpoint masks $M_s \in [0, 1]^{m \times k}$ and the object boxes $B\in \mathbb{R}^{k \times 6}$. 
Subsequently, the referent mask and box will be selected according to the referring scores $s\in \mathbb{R}^{k}$.

\begin{table*}[t!]
  \centering
\setlength{\tabcolsep}{6.5mm}  
\renewcommand{\arraystretch}{1.05} 
\begin{tabular}{l|cc|cc|cc}
\hline
\multicolumn{1}{c|}{} & \multicolumn{2}{c|}{Unique(19\%)} & \multicolumn{2}{c|}{Multiple(81\%)} & \multicolumn{2}{c}{Overall} \\
\multicolumn{1}{c|}{\multirow{-2}{*}{Method}}                      & 0.25             & 0.5            & 0.25              & 0.5             & 0.25          & 0.5         \\ \hline
ScanRefer \cite{scanrefer} & 67.64 & 49.19 & 32.06 & 21.26 & 38.97 & 26.10 \\
TGNN \cite{tgnn}           & 68.61 & 56.80 & 29.84 & 23.18 & 37.37 & 29.70 \\
3DJCG \cite{3djcg}         & 78.75 & 61.30 & 40.13 & 30.08 & 47.62 & 36.14 \\
D3Net \cite{d3net}         & -     & 70.35 & -     & 30.50 & -     & 37.87 \\
BUTD \cite{butd}           & 82.88 & 64.98 & 44.73 & 33.97 & 50.42 & 38.60 \\
EDA \cite{eda}             & 85.69 & 70.19 & 49.60 & 38.39 & 54.99 & 43.13 \\
\rowcolor{gray!10} \textbf{3DRefTR-HR (ours)} & \textbf{85.98} & \textbf{70.89} & \textbf{49.62} & 38.30 & \textbf{55.04} & \textbf{43.16} \\
\rowcolor{gray!10} \textbf{3DRefTR-SP (ours)} & \textbf{86.12} & \textbf{71.04} & \textbf{50.07} & \textbf{38.65} & \textbf{55.45} & \textbf{43.48} \\ \hline
\multicolumn{7}{l}{\textit{single-stage}} \\ 
\hline
3DSPS \cite{3dsps}         & 84.12 & 66.72 & 40.32 & 29.82 & 48.82 & 36.98 \\
BUTD \cite{butd}           & 81.47 & 61.24 & 44.20 & 32.81 & 49.76 & 37.05 \\
EDA \cite{eda}             & 86.40 & \textbf{69.42} & 48.11 & 36.82 & 53.83 & 41.70 \\
\rowcolor{gray!10} \textbf{3DRefTR-HR (ours)} & \textbf{86.75} & 67.94 & \textbf{48.73} & \textbf{37.79} & \textbf{54.40} & \textbf{42.29} \\
\rowcolor{gray!10} \textbf{3DRefTR-SP (ours)} & \textbf{86.40} & 68.01 & \textbf{48.82} & \textbf{37.83} & \textbf{54.43} & \textbf{42.33} \\ \hline
\end{tabular}
\caption{The 3D referring expression comprehension results on ScanRefer in terms of Acc@0.25IoU and Acc@0.5IoU. Our 3DRefTR surpasses existing state-of-the-art works in both two-stage and single-stage setting with a significant margin. }
\label{tab_3drec}
\end{table*}

\section{Experiments}
\subsection{DataSets} 
We assess our approach on 3D referring datasets, including ScanRefer \cite{scanrefer} and ReferIt3D \cite{referit3d}. These datasets derive from ScanNetv2 \cite{scannet}, featuring 1,513 detailed 3D indoor scene reconstructions. All datasets adhere to the official ScanNet splits.

\textbf{ScanRefer} offers 51,583 descriptions, averaging 13.81 objects and 64.48 descriptions per scene. Its evaluation metric is Acc@IoU, measuring the proportion of descriptions where the predicted box and ground truth overlap with an IoU greater than 0.25 and 0.5. Descriptions are categorized into "unique" if the object is the sole representative of its class in the scene, or "multiple" otherwise.

\textbf{ReferIt3D} includes two subsets: Sr3D and Nr3D. Sr3D encompasses 83,572 template-generated expressions, while Nr3D houses 41,503 human-generated expressions. Instead of presenting the full scene, this dataset offers segmented point clouds for individual objects. The primary evaluation metric for ReferIt3D is accuracy, gauging if the model aptly identifies the target object.

\subsection{Implementation Details}
Our model is trained with the AdamW optimizer and a batch size of 12 on four NVIDIA V100 GPUs for 100 epochs.
We freeze the text backbone Robeta-base following the settings in \cite{eda}, while the rest of the network is trainable.
The initial learning rates of the PointNet++ backbone and the rest of the model are empirically set to 2e-3 and 2e-4, respectively. 
We apply learning rate decay at epochs 50 and 75 with a rate of 0.1. 
For all the datasets, we use xyz coordinates and RGB values as the input, and the number of visual tokens $V$ and query embeddings $Q$ are empirically set to 1024 and 256, respectively. 
The sample number and radius of the ball query are set to 2 and 0.2, respectively. 
The number of decoder layers $L$ is set to 6.
The balancing factors are set default as $\alpha_1$ = 1.0 / ($L$ + 1), $\alpha_2$ = 10.0, $\alpha_3$ = 2.0, $\alpha_4$ = 8.0, $\beta_1$ = 5.0, $\beta_2$ = 1.0, $\beta_3$ = 1.0, $\beta_4$ = 0.5 and $\beta_5$ = 0.5 for the ScanRefer dataset. For the Nr3D/Sr3D dataset, $\beta_4$ and $\beta_5$ are adjusted to be 1.0 and 1.0, respectively.

\begin{table}[t!]
  \centering
\setlength{\tabcolsep}{5.5mm}
\begin{tabular}{l|cc}
\hline
Method                     & Sr3D          & Nr3D          \\ \hline
ReferIt3D \cite{referit3d}                 & 39.8          & 35.6          \\
TGNN \cite{tgnn}                                        & 45.0                     & 37.3                     \\
3DSPS \cite{3dsps}                                       & 62.6                     & 51.5                     \\
BUTD \cite{butd}                                        & 65.6                     & 49.1                     \\
EDA \cite{eda}                                         & 68.1                     & 52.1                     \\
\rowcolor{gray!10}\textbf{3DRefTR-SP (ours)}                   & \textbf{68.5}            & \textbf{52.6}            \\ \hline
\end{tabular}
    \caption{The 3DREC performance on SR3D/NR3D datasets by Acc@0.25IoU as the metric.  }
    \label{tab_3drec_it}
\end{table}

\subsection{Quantitative Comparison}
In this subsection, we perform a qualitative comparison on ScanRefer \cite{scanrefer} and Referit3D \cite{referit3d}. 

\textbf{3DRES. }
To the best of our knowledge, TGNN \cite{tgnn} is the only relevant study for the 3DRES task.
To ensure a more fair comparison, we established two baseline methods, namely 3DRESTR and EDA-box2mask, based on the state-of-the-art 3DREC approach EDA \cite{eda}. 
The 3DRESTR shares a nearly identical network structure with EDA, with the sole distinction being the substitution of the box regression branch in the object decoder with a mask branch. Meanwhile, the EDA-box2mask converts the box output from the EDA model into a mask, i.e., filtering points in the 3D scene using the outputted 3D bounding box. 
As reflected in Tab. \ref{tab_3dres}, our proposed 3DRefTR significantly outperforms these baseline methods across both two-stage and single-stage settings. 
Specifically, in terms of mIoU, the two-stage 3DRefTR-SP surpasses TGNN, 3DRESTR, and EDA-box2mask by margins of 12.96\%, 12.02\%, and 5.73\% respectively. 
The performance of single-stage 3DRefTR-SP is comparably impressive. Notably, although 3DRefTR-HR shows superior segmentation performance over 3DRefTR-SP, the marginal gains do not seem compelling when weighed against its much higher computational demands. 

\textbf{3DREC.}
To evaluate the 3DREC performance, we compare 3DRefTR with several existing 3DREC works on ScanRefer and ReferIt3D, involving 
the state-of-the-art DETR-like methods EDA \cite{eda} and BUTD \cite{butd}, 
the single-stage model 3D-SPS \cite{3dsps}, 
the segmentation-based two-stage method TGNN (transform the predicted mask into the box) \cite{tgnn}, 
among others.
As evidenced in Tab. \ref{tab_3drec} and Tab. \ref{tab_3drec_it}, 3DRefTR consistently exhibits superior performance over these prevailing methods in both two-stage and single-stage settings. 
This evidence strongly suggests that the newly incorporated mask prediction branch has indeed augmented the performance of the 3DREC task. The commendable results can be largely attributed to the well-constructed design of our unified 3DVG framework.


\begin{table}[t!]
  \centering
\begin{tabular}{l|cc}
\hline
Training Scheme           & 3DRES          & 3DREC          \\ \hline
only 3DREC                & -              & 54.99          \\
only 3DRES                & 28.74          & -              \\ \hline
joint scratch training    & \textbf{41.39} & 54.61          \\
3DRES then joint          & 41.21          & 54.70          \\
3DREC then joint          & 40.65          & 54.02          \\
3DREC frozen              & 40.20          & 54.99          \\
3DREC then joint (two lr) & 40.76          & \textbf{55.45} \\ \hline
\end{tabular}
\caption{Performance on Scanrefer affected by different training schemes. 'two lr' means jointly training 3DRES and 3DREC components with different learning rates.}
\label{tab_traing}
\end{table}

\begin{figure*}[h!]
  \centering 
  \includegraphics[width=1.0\linewidth]{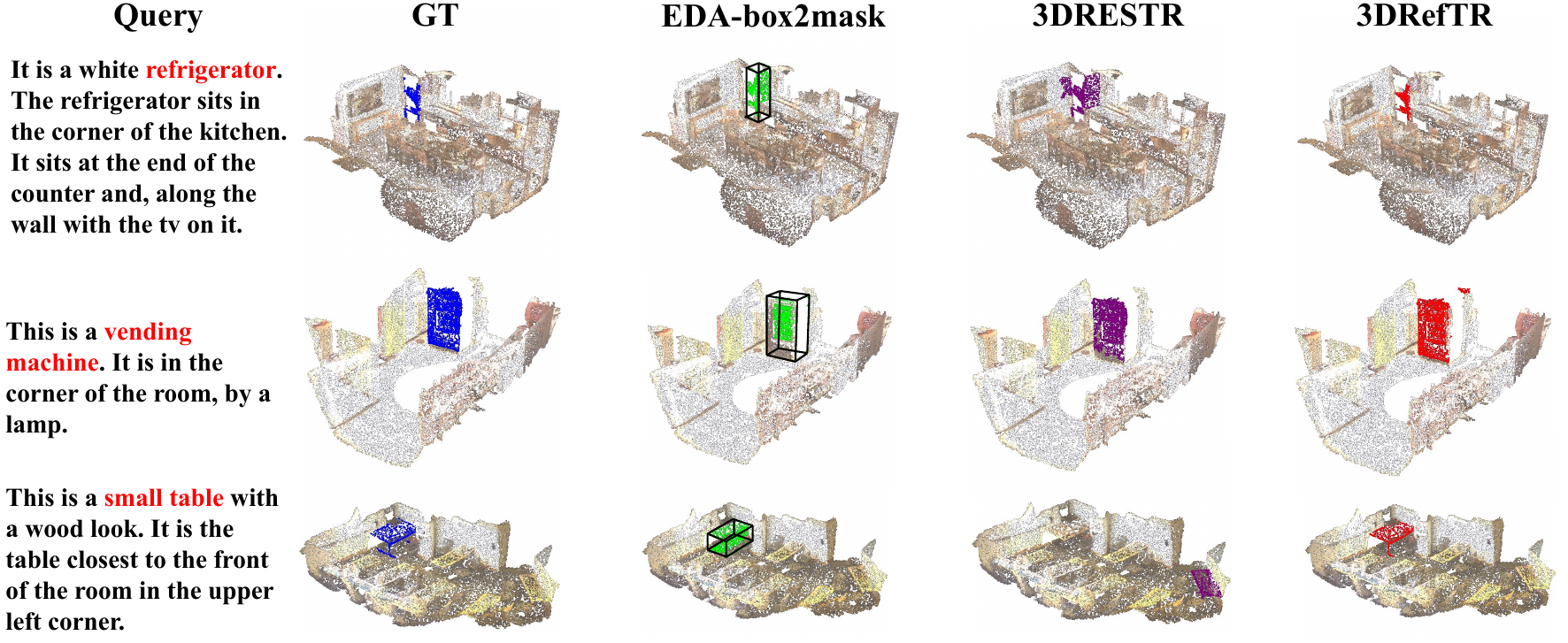}
  \caption{Visualization of the 3D refering expression segmentation results. EDA-box2mask has difficulties with mask boundaries. 3DRESTR exhibits weak localization. However, 3DRefTR-SP excels in both localization and mask refinement.  }
  \label{fig_vis} 
\end{figure*}

\subsection{Ablation Study}\label{sec_ablation}
In this subsection, we delve into the influence of varying configurations on the performance of 3DRefTR. 
The experiments in this subsection are conducted on Scanrefer
and metrics adopted for 3DREC and 3DRES performance are Acc@0.25IoU and mIoU, repectively.
Due to space constraints in the main text, we have described additional ablation experiments in the supplementary materials. 

\textbf{Joint Training Scheme. }
We explore the impacts of different joint training schemes on task performance. 
Five training schemes were investigated for 3DRefTR: 
1) jointly training from scratch, 
2) loading the pretrained 3DRES model and jointly training 3DRES and 3DREC, 
3) loading the pretrained 3DREC model and jointly training 3DRES and 3DREC, 
4) freezing the pretrained 3DREC component and training the Superpoint Mask Branch alone, 
5) loading the pretrained 3DREC model and jointly training 3DRES and 3DREC with different learning rates ($2e-4$ and $2e-6$). 
As displayed in Tab. \ref{tab_traing}, 
The 1st, 2nd, and 3rd scheme bring about significant 3DRES performance gain but cause a slight performance drop on 3DREC. 
This indicates that while 3DREC significantly boosts 3DRES by providing precise localisation cues, the enhancement of 3DREC by 3DRES is more modest. 
The fine-grained visual-language alignment contributed by 3DRES yields smaller gains in 3DREC, which is easily overshadowed by losses incurred from the diverging optimization objectives inherent in each task. 
This inspires us to apply a smaller learning rate on the 3DREC components (the 5th scheme) for better trade-off. 
The 5th scheme with different learning rates for 3DRES and 3DREC demonstrates the most optimal trade-off.

\textbf{Discussion on 3DRESTR.}
To determine the causes of 3DRESTR's suboptimal performance, we systematically carried out step-by-step experiments. Table \ref{tab_3drestr} reveals a progressive enhancement in performance as box-regression components were added to 3DRESTR. 
This highlights the substantial improvement that the localization information provided by the 3DREC task contributes to the 3DRES task.
Specifically, the addition of a single box regression layer (proposal) led to a significant performance boost of +3.4\%. Incorporating all seven layers (proposal + multiple) brought performance closer (+12.43\%) to the optimal structure. In contrast, the contribution of box positional embedding was more modest (+0.22\%).

\textbf{Hyperparameters of Superpoint Query.}
To acquire superpoint embeddings, the ball query method \cite{pointnet2} is used in Superpoint Mask Branch. 
Specifically, using each superpoint as the center, a ball  with radius $R$ is defined in the coordinate space. 
Features of $N_{sample}$ points are then sampled within this ball, with superpoint features subsequently obtained through max pooling. 
To investigate the impact of Superpoint Query sampling range on model efficacy, the values of $R$ and $N_{sample}$ were varied. 
Results are presented in Table \ref{tab_ball}, which indicates that the model achieves trade-off performance with $R$ at 0.2 and $N_{sample}$ at 2.

\begin{table}[t]
  \centering
\begin{tabular}{l|c}
\hline
\multicolumn{1}{l|}{Model} & mIoU \\ \hline
3DRESTR                           &   28.74     \\
3DRESTR + proposal                &   32.14     \\
3DRESTR + proposal + multiple                &   41.17     \\
3DRESTR + proposal + multiple + position     &   \textbf{41.39}     \\ \hline
\end{tabular}
\caption{Progressive exploration of 3DRESTR performance on ScanRefer dataset. "proposal" refers to the integration of a proposal box head preceding the object decoder, "multiple" denotes the addition of a box regression head after each layer within the object decoder, and "position" signifies the incorporation of box positional embedding within the object decoder. The last row is equivalent to 3DRefTR-SP. }
\label{tab_3drestr}
\end{table}

\begin{table}[t]
  \centering
\setlength{\tabcolsep}{4.0mm}
\begin{tabular}{cc|cc}
\hline
$R$  &  $N_{sample}$  & 3DRES          & 3DREC          \\ \hline
\multirow{3}{*}{0.2}     &  2  & 40.76 &   \textbf{55.45}       \\
                          &  4  & \textbf{41.38}          & 55.29          \\
                          &  8  & 40.97          & 55.22          \\ \hline
\multirow{3}{*}{0.4}        &  2  & 35.48 & 55.15          \\
                              &  4  &     37.83      &     55.19      \\
                              &  8  &      32.50     &     55.10      \\ \hline
\multirow{3}{*}{0.6}        &  2  & 29.55 &    55.04      \\
                          &  4  &     33.28      &     55.20      \\
                          &  8  &     36.35      &    55.22       \\ \hline
\end{tabular}
\caption{Performance on Scanrefer affected by different hyperparameters of Ball Query. The metrics for 3DREC and 3DRES are Acc@0.25IoU and mIoU.}
\label{tab_ball}
\end{table}

\subsection{Qualitative Comparison}
The visualization of the results for 3D Referring Expression Segmentation is presented in Fig. \ref{fig_vis}. 
In this figure, the first column depicts the referring expressions, the second column showcases the ground truth segmentation mask, and the subsequent columns present the mask predictions of EDA-box2mask, 3DRESTR, and 3DRefTR-SP, respectively.
EDA-box2mask displays commendable localization capabilities. 
However, the points filtered out by its box often encounter challenges in handling the mask boundaries. 
As a consequence, there are frequent instances of incomplete masks or residues from other objects. 
3DRESTR, on the other hand, exhibits prowess in precisely managing mask boundaries. 
Nevertheless, its localization ability is found wanting, as evidenced in the third row, resulting in the lowest mIoU score.
Contrastingly, our proposed 3DRefTR-SP excels in both precise localization and impeccable handling of mask boundaries. 

\section{Conclusions}
This work introduced the 3D Referring Transformer, 3DRefTR, an innovative approach that concurrently addresses 3DREC and 3DRES. 
Through the strategic integration of a superpoint-based mask branch, the model not only advances the state-of-the-art in high-resolution segmentation but also achieves a refined alignment between linguistic and visual representations, further bolstering the performance of 3DREC.
A significant innovation in our framework is the exploitation of heterogeneous CPU-GPU parallelism. 
By leveraging this parallelism, 3DRefTR efficiently harnesses the capabilities of both processing units, allowing the CPU to generate superpoints concurrently while the GPU produces visual tokens. 
This parallel processing mitigates the computational challenges traditionally associated with high-resolution visual tasks, ensuring efficient and scalable model operation.
To conclude, the 3DRefTR framework marks a step forward in 3D visual grounding. 
Its ability to harmoniously unify 3DREC and 3DRES tasks, backed by computational efficiency and good performance, underscores its potential in advancing the field of 3D scene comprehension. 

{
    \small
    \bibliographystyle{ieeenat_fullname}
    \bibliography{egbib}
}


\end{document}